\documentclass[11pt,a4paper]{article}
\usepackage[hyperref]{naaclhlt2019}
\usepackage{times}
\usepackage{latexsym}
\usepackage{url}

\aclfinalcopy 
\def\aclpaperid{817} 

\usepackage{booktabs} 
\usepackage{subcaption}
\usepackage{graphicx}
\usepackage{amsmath}
\allowdisplaybreaks
\usepackage{amssymb}
\usepackage[inline]{enumitem}
\usepackage{makecell}
\usepackage{multirow}
\usepackage{xcolor}

\title{A Causality-Guided Prediction of the TED Talk Ratings from the Speech-Transcripts using Neural Networks}

\author{Md Iftekhar Tanveer$^1$, Md Kamrul Hassan$^2$, Daniel Gildea$^3$, M. Ehsan Hoque$^4$ \\
    University of Rochester\\
    {\tt \{$^1$itanveer,$^2$mhasan8,$^3$gildea,$^4$mehoque\}@cs.rochester.edu}}

\date{}

\begin{document}
\maketitle
\begin{abstract}
Automated prediction of public speaking performance enables novel systems for tutoring public speaking skills. We use the largest open repository---TED Talks---to predict the ratings provided by the online viewers. The dataset contains over 2200 talk transcripts and the associated meta information including over 5.5 million ratings from spontaneous visitors to the website. We carefully removed the bias present in the dataset (e.g., the speakers' reputations, popularity gained by publicity, etc.) by modeling the data generating process using a causal diagram. We use a word sequence based recurrent architecture and a dependency tree based recursive architecture as the neural networks for predicting the TED talk ratings. Our neural network models can predict the ratings with an average F-score of 0.77 which largely outperforms the competitive baseline method.
\end{abstract}



\section{Introduction}
While the demand for physical and manual labor is gradually declining, there is a growing need for a workforce with soft skills. Which soft skill do you think would be the most valuable in your daily life? According to an article in Forbes~\cite{Gallo2014a}, 70\% of employed Americans agree that public speaking skills are critical to their success at work. Yet, it is one of the most dreaded acts. Many people rate the fear of public speaking even higher than the fear of death~\cite{Wallechinsky2005}. To alleviate the situation, several automated systems are now available that can quantify behavioral data for participants to reflect on~\cite{Fung2015}. Predicting the viewers' ratings from the speech transcripts would enable these systems to generate feedback on the potential audience behavior.

Predicting human behavior, however, is challenging due to its huge variability and the way the variables interact with each other. Running Randomized Control Trials (RCT) to decouple each variable is not always feasible and also expensive. It is possible to collect a large amount of observational data due to the advent of content sharing platforms such as YouTube, Massive Open Online Courses (MOOC), or \url{ted.com}. However, the uncontrolled variables in the observational dataset always keep a possibility of incorporating the effects of the ``data bias'' into the prediction model. Recently, the problems of using biased datasets are becoming apparent. \citet{buolamwini2018gender} showed that the error rates in the commercial face-detectors for the dark-skinned females are $43$ times higher than the light-skinned males due to the bias in the training dataset. The unfortunate incident of Google's photo app tagging African-American people as ``Gorilla''~\cite{guynn_2015} also highlights the severity of this issue.

We address the data bias issue as much as possible by carefully analyzing the relationships of different variables in the data generating process. We use a Causal Diagram~\cite{Pearl2018,Pearl2009} to analyze and remove the effects of the data bias (e.g., the speakers' reputations, popularity gained by publicity, etc.) in our prediction model. In order to make the prediction model less biased to the speakers' race and gender, we confine our analysis to the transcripts only. Besides, we normalize the ratings to remove the effects of the unwanted variables such as the speakers' reputations, publicity, contemporary hot topics, etc.

For our analysis, we curate an observational dataset of public speech transcripts and other meta-data collected from the \url{ted.com} website. This website contains a large collection of high-quality public speeches that are freely available to watch, share, rate, and comment on. Every day, numerous people watch and annotate their perceptions about the talks. Our dataset contains $2231$ public speech transcripts and over $5$ million ratings from the spontaneous viewers of the talks. The viewers annotate each talk by 14 different labels---\emph{Beautiful}, \emph{Confusing}, \emph{Courageous}, \emph{Fascinating}, \emph{Funny}, \emph{Informative}, \emph{Ingenious}, \emph{Inspiring}, \emph{Jaw-Dropping}, \emph{Long-winded}, \emph{Obnoxious}, \emph{OK}, \emph{Persuasive}, and \emph{Unconvincing}.

We use two neural network architectures in the prediction task. In the first architecture, we use LSTM~\cite{Hochreiter1997} for a sequential input of the words within the sentences of the transcripts. In the second architecture, we use TreeLSTM~\cite{Tai2015} to represent the input sentences in the form of a dependency tree. Our experiments show that the dependency tree-based model can predict the TED talk ratings with slightly higher performance (average F-score 0.77) than the word sequence model (average F-score 0.76). To the best of our knowledge, this is the best performance in the literature on predicting the TED talk ratings. We compare the performances of these two models with a baseline of classical machine learning techniques using hand-engineered features. We find that the neural networks largely outperform the classical methods. We believe this gain in performance is achieved by the networks' ability to capture better the natural relationship of the words (as compared to the hand engineered feature selection approach in the baseline methods) and the correlations among different rating labels.

\section{Background Research}
In this section, we describe a few relevant prior arts on behavioral prediction.

\subsection{Predicting Human Behavior}
An example of human behavioral prediction research is to \emph{automatically grade essays}, which has a long history~\cite{valenti2003overview}. Recently, the use of deep neural network based solutions~\cite{alikaniotis2016automatic,taghipour2016neural} are becoming popular in this field. \citet{farag2018neural} proposed an adversarial approach for their task. \citet{jin2018tdnn} proposed a two-stage deep neural network based solution. Predicting \emph{helpfulness}~\cite{martin2014prediction,yang2015semantic,liu2017using,chen2018cross} in the online reviews is another example of predicting human behavior. \citet{bertero2016long} proposed a combination of Convolutional Neural Network (CNN) and Long Short-Term Memory (LSTM) based framework to \emph{predict humor} in the dialogues. Their method achieved an 8\% improvement over a Conditional Random Field baseline. \citet{jaech2016phonological} analyzed the performance of \emph{phonological pun detection} using various natural language processing techniques. In general, behavioral prediction encompasses numerous areas such as predicting \emph{outcomes in job interviews}~\cite{Naim2016}, \emph{hirability}~\cite{Nguyen2016}, \emph{presentation performance}~\cite{Tanveer2015,Chen2017a,Tanveer2018} etc. However, the practice of explicitly modeling the data generating process is relatively uncommon. In this paper, we expand the prior work by explicitly modeling the data generating process in order to remove the data bias.

\subsection{Predicting the TED Talk Performance} There is a limited amount of work on predicting the TED talk ratings. In most cases, TED talk performances are analyzed through introspection~\cite{Gallo2014,Bull2016,sugimoto2013scientists,tsou2014community,drasovean2015evaluative}. 

\citet{Chen2017} analyzed the TED Talks for humor detection. \citet{Liu2017} analyzed the transcripts of the TED talks to predict audience engagement in the form of applause. \citet{Haider2017} predicted user interest (engaging vs. non-engaging) from high-level visual features (e.g., camera angles) and audience applause. \citet{Pappas:2013:SAU:2484028.2484116} proposed a sentiment-aware nearest neighbor model for a multimedia recommendation over the TED talks. \citet{weninger2013words} predicted the TED talk ratings from the linguistic features of the transcripts. This work is most similar to ours. However, we are proposing a new prediction framework using the Neural Networks.

\section{Dataset}\label{sec:DatContents}
The data for this study was gathered from the \url{ted.com} website on November 15, 2017. We removed the talks published six months before the crawling date to make sure each talk has enough ratings for a robust analysis. More specifically, we filtered any talk that--- \begin{enumerate*} \item was published less than $6$ months prior to the crawling date, \item contained any of the following keywords: live music, dance, music, performance, entertainment, or, \item contained less than 450 words in the transcript\end{enumerate*}. This filtering left a total of 2231 talks in the dataset.

\begin{figure}
\centering
\includegraphics[width=1\linewidth]{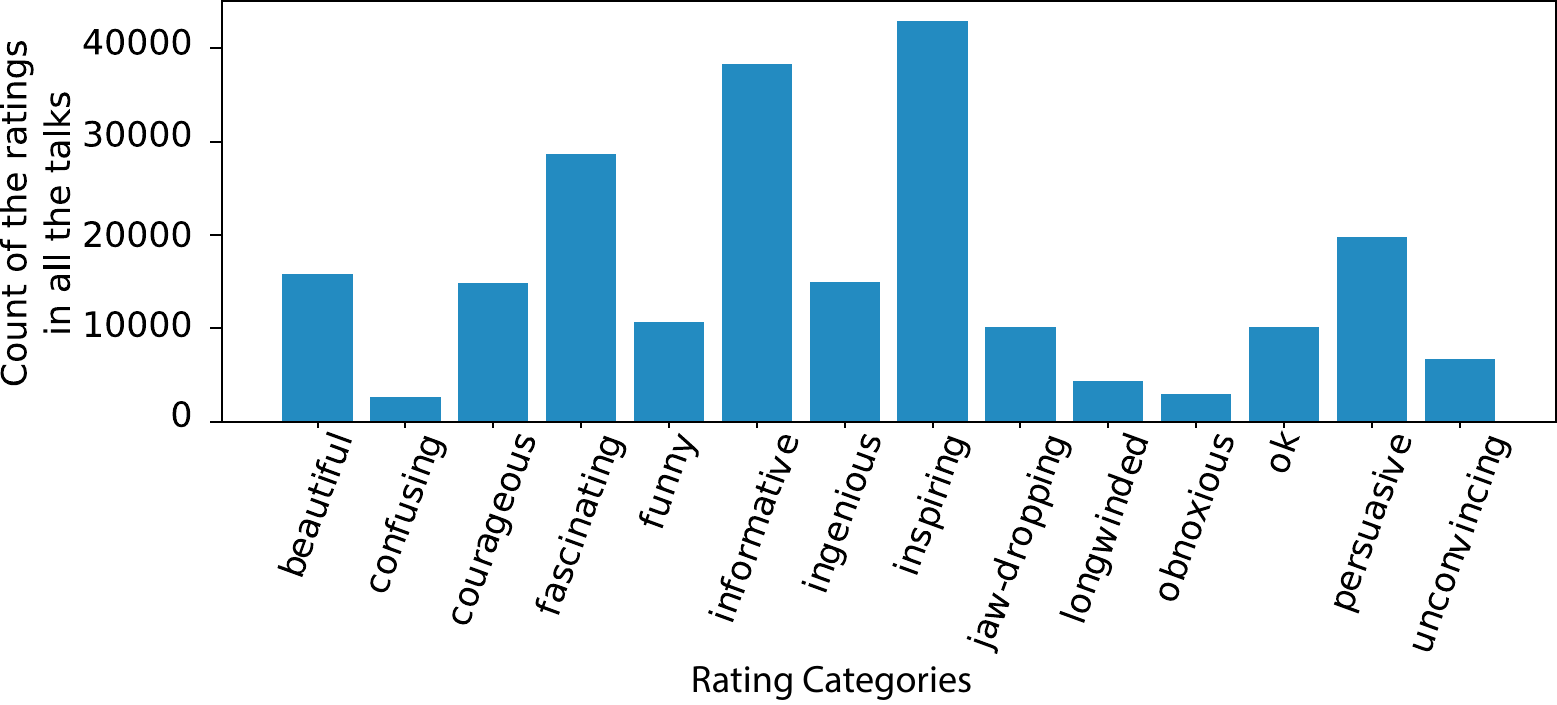}
\caption{Counts of all the 14 different rating categories (labels) in the dataset}
\label{fig:rating_counts}
\end{figure}

We collected the manual transcriptions, and the total view counts for each video. We also collected the ``ratings'' which are the counts of the viewer-annotated labels. The viewers can annotate a talk from a selection of 14 different labels provided in the website. The labels are not mutually exclusive. Viewers can choose at most $3$ labels for each talk. If only one label is chosen, it is counted $3$ times. We count the total number of annotations under each label as shown in Figure~\ref{fig:rating_counts}. The ratings are treated as the ground truth about the audience perception. A summary of the dataset characteristics is shown in Table~\ref{tab:datasize}.

\begin{table}
  \centering
  \begin{tabular}{ll}
    \toprule
    \textbf{Property}& \textbf{Quantity}\\
    \midrule
\textbf{Number of talks} & 2,231\\
\textbf{Total length of all talks} & 513.49 Hours\\
\textbf{Total number of ratings} & 5,574,444\\
\textbf{Average ratings per talk} & 2498.6\\
\textbf{Minimum ratings per talk} & 88\\
\textbf{Total word count} & 5,489,628\\
\textbf{Total sentence count} & 295,338\\
  \bottomrule
\end{tabular}
  \caption{Dataset Properties}
  \label{tab:datasize}
\end{table}

\section{Modeling the Data Generating Process}
In order to analyze and remove any bias present in the dataset, we model the data generating process using a Causal Diagram. For a delightful understanding of the importance of this step, please refer to Chapter 6 of \citet{Pearl2018}.

\begin{figure}
\centering
\includegraphics[width=1\linewidth]{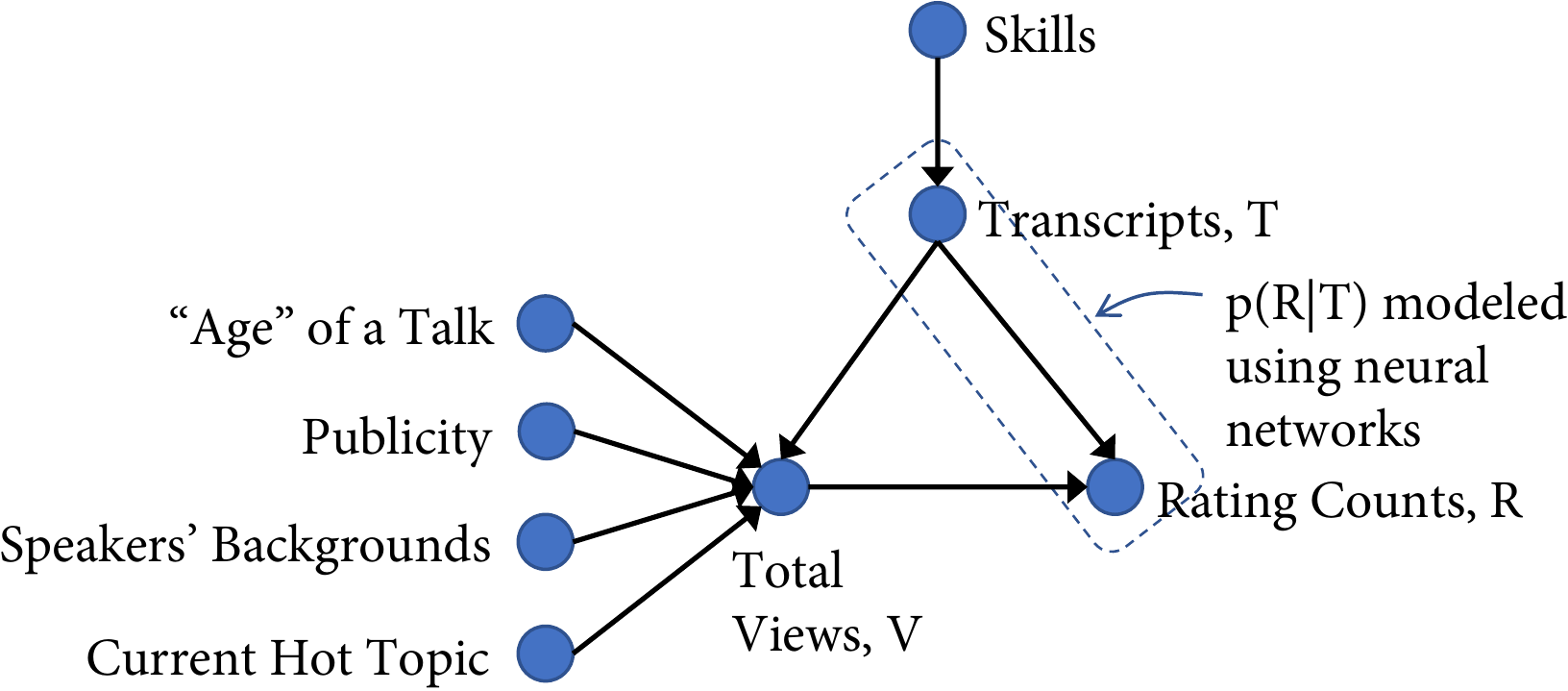}
\caption{Causal Diagram of the Data Generating Process of TED Talks}
\label{fig:data-gen-proc}
\end{figure}

The (assumed) causal diagram of the TED talk data generating process is shown in Figure~\ref{fig:data-gen-proc}.
We know that the popularity (i.e. \emph{Total Views}, $V$) depends on the speech contents (i.e. \emph{Transcripts}). Therefore, the \emph{Transcripts} ($T$) cause the \emph{Total Views} ($V$) to change and thus we draw an arrow from \emph{Transcripts} to \emph{Total Views}. Although we know that the popularity also depends on the nonverbal contents (e.g. prosody, facial expressions), we remove those modalities from our prediction system for eliminating any gender or racial bias. \emph{Transcripts} also cause the distribution of the \emph{Rating Counts} ($R$) to change. Now, the \emph{Total Views} also cause the \emph{Rating Counts} to change; because, the more the people watch a specific talk, the high the rating counts. We can safely assume this arrow from \emph{Total Views} to \emph{Rating Counts} models a linear relationship.

The causal relationships so far reveal that the \emph{Total Views}, $V$ act as a ``mediator''~\cite{Pearl2018} between $T$ and $R$ and thus helps our prediction. However, it is easy to see that $V$ is affected by various biases present in the dataset. For example, the longer a TED talk remains on the web, the more views it gets. Therefore, the \emph{``Age'' of a talk} causes the \emph{Total Views} to change. We can imagine many other variables (e.g., how much the talk is publicized, the speakers' reputations) can affect the \emph{Total Views}. We, however, do not want these variables to affect our prediction. Fortunately, all these variables can affect the \emph{Rating Counts} only through the \emph{Total Views}---because the viewers must arrive into the page in order to annotate the ratings. Therefore, we can remove the effects of the unwanted variables by removing the effects of the \emph{Total Views} from the \emph{Rating Counts} with the help of the linearity assumption mentioned before. We normalize the rating counts of each talk as in the following equation:
\begin{equation}\label{eq:scaled_score}
r_{i,\text{scaled}} = \frac{r_i}{\sum_{j=1}^{14}{r_j}} 
\end{equation}
Where $r_i$ represents the count of the $i^{\text{th}}$ label in a talk. Let us assume that in a talk, $f_i$ fractions of the total viewers annotate for the rating category $i$. Then the scaled rating, $r_{i,\text{scaled}}$ becomes $\frac{f_iV}{\sum_j{f_jV}}=\frac{Vf_i}{V\sum_j{f_j}}$. Notice that the \emph{Total Views}, $V$ gets canceled from the numerator and the denominator for each talk. This process successfully removes the effect of $V$ as evident in Table~\ref{tab:corrcoef}. Scaling the rating counts removes the effects of \emph{Total Views} by reducing the average correlation from $0.56$ to $-0.03$. This process also removes the effect of the \emph{Age of the Talks} by reducing the average correlation from $0.15$ to $0.06$. Therefore, removing $V$ reduces the effect of the \emph{Age of the Talks} in the ratings. It should work similarly for the other unwanted variables as well.
\begin{table}
\centering
\begin{tabular}{lrrrr} 
\toprule
                       & \multicolumn{2}{l}{\textbf{\makecell{Total Views}}} & \multicolumn{2}{l}{\textbf{\makecell{Age of Talks}}}  \\
                       & \textbf{noscale} & \textbf{scale}        & \textbf{noscale} & \textbf{scale}          \\ 
\midrule
\textbf{Beaut.}     & 0.52             & 0.01                  & 0.03             & -0.14                   \\
\textbf{Conf.}     & 0.39             & -0.12                 & 0.27             & 0.20                    \\
\textbf{Cour.}    & 0.52             & -0.003                & 0.01             & 0.15                    \\
\textbf{Fasc.}   & 0.78             & 0.05                  & 0.15             & 0.06                    \\
\textbf{Funny}         & 0.57             & 0.14                  & 0.10             & 0.10                    \\
\textbf{Info.}   & 0.76             & -0.08                 & 0.07             & -0.19                   \\
\textbf{Ingen.}      & 0.59             & -0.06                 & 0.18             & 0.10                    \\
\textbf{Insp.}     & 0.79             & 0.1                   & 0.05             & -0.15                   \\
\textbf{Jaw-Dr.}  & 0.51             & 0.1                   & 0.18             & 0.23                    \\
\textbf{Long.}    & 0.44             & -0.17                 & 0.36             & 0.31                    \\
\textbf{Obnox.}     & 0.27             & -0.11                 & 0.19             & 0.17                    \\
\textbf{OK}            & 0.72             & -0.16                 & 0.21             & 0.14                    \\
\textbf{Pers.}    & 0.72             & -0.01                 & 0.12             & 0.02                    \\
\textbf{Unconv.}  & 0.29             & -0.14                 & 0.18             & 0.15                    \\ 
\midrule
\textbf{Avg.}       & 0.56             & -0.03                 & 0.15             & 0.06                    \\
\bottomrule
\end{tabular}
\caption{Correlation coefficients of each category of the ratings with the \emph{Total Views} and the \emph{``Age'' of Talks}}
\label{tab:corrcoef}
\end{table}

We binarize the scaled ratings by thresholding over the median value which results in a $0$ and $1$ class for each category of the ratings. The label $1$ indicates having a rating higher than the median value. We model $p(R_\text{scaled}|T)$ using neural networks as discussed in the following sections.  $R_\text{scaled}$ refers to the scaled and binarized ratings.

\section{Network Architectures}
We implement two neural networks to model $p(R_\text{scaled}|T)$. Architectures of these networks are described below.

\subsection{Word Sequence Model}
\begin{figure}
\centering
\includegraphics[width=1\linewidth]{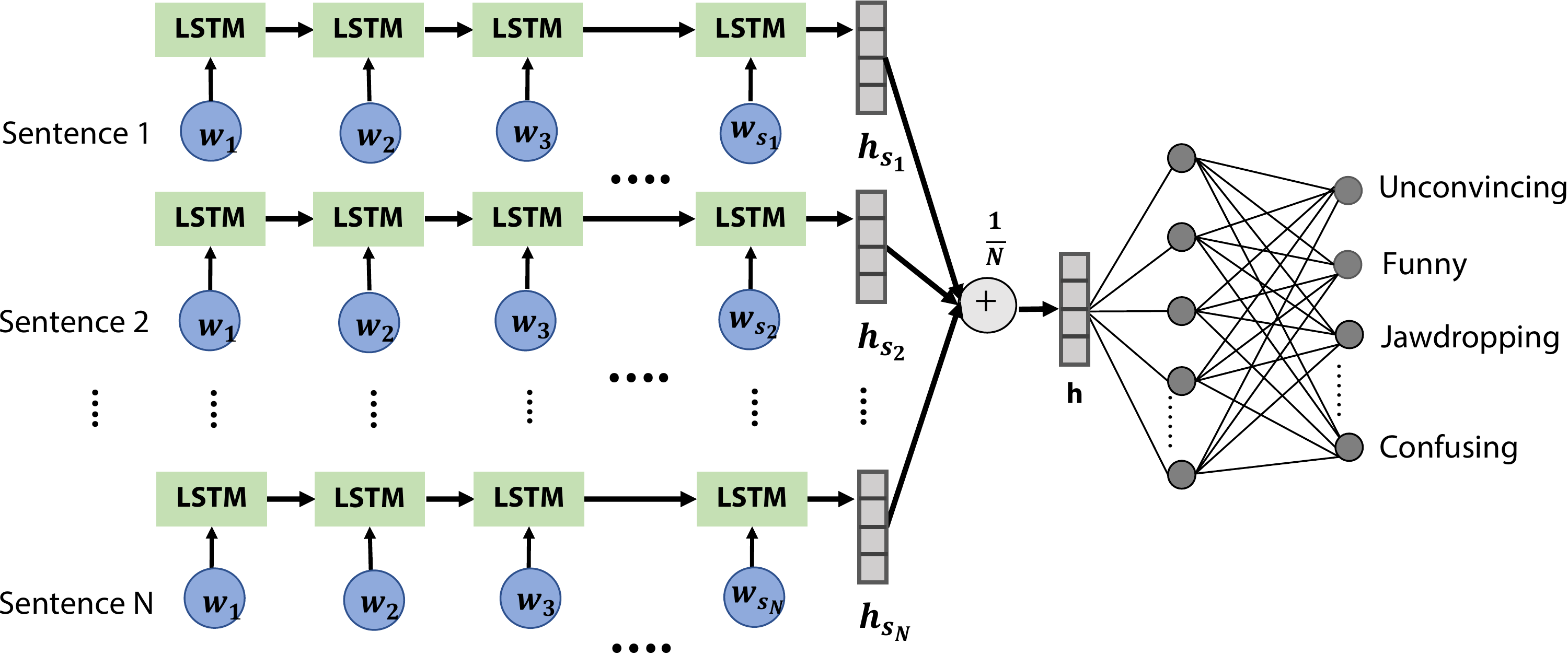}
\caption{An illustration of the Word Sequence Model}
\label{fig:model_word_seq}
\end{figure}

We use a Long Short Term Memory (LSTM) \cite{Hochreiter1997} neural network to model the word-sequences in the transcripts. However, the transcripts have around $2460$ words on average. It is difficult to model such a long chain even with an LSTM due to the vanishing/exploding gradient problem. We, therefore, adopt a ``Bag-of-Sentences'' model where we model each sentence using the LSTM and average the outputs for predicting the scaled and binarized rating counts. A pictorial illustration of this model is shown in Figure~\ref{fig:model_word_seq}.
Each sentence, $s_j$ in the transcript is represented by a sequence of word-vectors\footnote{In this paper, we represent the column vectors as lowercase boldface letters; matrices or higher dimensional tensors as uppercase boldface letters and scalars as lowercase regular letters. We use a prime symbol ($'$) to represent the transpose operation.}, $\mathbf{w}_1,\mathbf{w}_2,\mathbf{w}_3,\dots,\mathbf{w}_{n_j}$. Here, each $\mathbf{w}$ represents the pre-trained, 300-dimensional GLOVE word vectors~\cite{pennington2014glove}. We use an LSTM to obtain an embedding vector, $\mathbf{h}_{s_j}$, for the $j^{\text{th}}$ sentence in the talk transcript. These vectors ($\mathbf{h}_{s_j}$) are averaged and passed through a feed-forward network to produce a 14-dimensional output vector corresponding to the categories of the ratings. An element-wise sigmoid ($\sigma(x) = \frac{1}{1+e^{-x}}$) activation function is applied to the output vector. The mathematical description of the model is as follows:
\begin{align}
&\mathbf{i}_t =\sigma(\mathbf{U}_i\mathbf{w}_t+\mathbf{V}_i\mathbf{h}_{t-1} + \mathbf{b}_i)\label{eq:lstm_first}\\
&\mathbf{f}_t =\sigma(\mathbf{U}_f\mathbf{w}_t+\mathbf{V}_f\mathbf{h}_{t-1} + \mathbf{b}_f)\\
&\mathbf{u}_t =\tanh(\mathbf{U}_u\mathbf{w}_t+\mathbf{V}_u\mathbf{h}_{t-1} + \mathbf{b}_u)\\
&\mathbf{o}_t =\sigma(\mathbf{U}_o\mathbf{w}_t+\mathbf{V}_o\mathbf{h}_{t-1} + \mathbf{b}_o)\label{eq:Vmatend}\\
&\mathbf{c}_t =\mathbf{f}_t\odot \mathbf{c}_{t-1} + \mathbf{i}_t\odot\mathbf{u}_t\\
&\mathbf{h}_t = \mathbf{o}_t\odot\tanh(\mathbf{c}_t)\label{eq:lstm_last} \\
&\mathbf{h}_{s_j} = \mathbf{h}_{n_j}\\
&\mathbf{h} = \frac{1}{N}\sum_{j=1}^N\mathbf{h}_{s_j}\\
&\mathbf{r} = \sigma(\mathbf{W}\mathbf{h} + \mathbf{b}_r)
\end{align}
Here, equations~\eqref{eq:lstm_first} to \eqref{eq:lstm_last} represent the definitive characteristics of LSTM. The vectors $\mathbf{i}_t$, $\mathbf{f}_t$ and $\mathbf{o}_t$ are the input, forget, and output gates (at the $t^{\text{th}}$ position), respectively; $\mathbf{c}_t$ and $\mathbf{h}_t$ represent the memory cell and the hidden states of the LSTM. The $\odot$ notation represents the Hadamard (element-wise) product between two vectors. We chose the sentence embeddings to have 128 dimensions; therefore, the dimensions of the transformation matrices $\mathbf{U}$'s, $\mathbf{V}$'s, $\mathbf{W}$'s are $128\times300$, $128\times128$, and $128\times14$ respectively. $\mathbf{U}$'s, $\mathbf{V}$'s, $\mathbf{b}$'s and $\mathbf{W}$ are the free parameters of the network which are learned through back-propagation. The output vector $\mathbf{r}$ represents $p(R_\text{scaled}|T)$. For the $j^{\text{th}}$ sentence, the index $t$ varies from $1$ to $n_j$, where $n_j$ is the number of words in the $j^\text{th}$ sentence. $N$ represents the total number of the sentences in the transcript. We use zero vectors to initialize the memory cell ($\mathbf{c}_0$) and the hidden state ($\mathbf{h}_0$) and as any out-of-vocabulary word vector.

\subsection{Dependency Tree-based Model}\label{sec:deptree_model}
\begin{figure}
\centering
\includegraphics[width=1\linewidth]{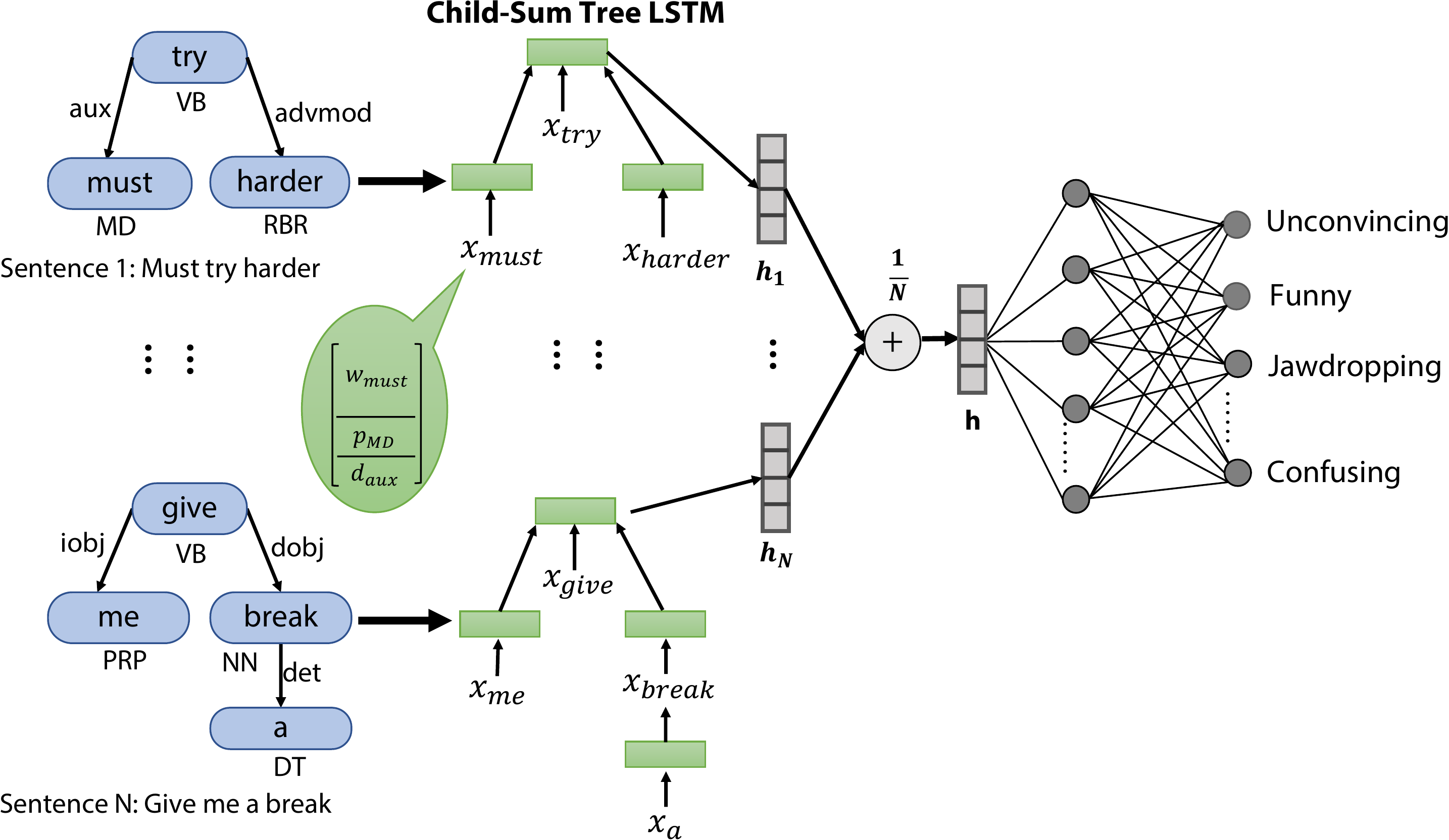}
\caption{An illustration of the Dependency Tree-based Model}
\label{fig:model_dep_tree}
\end{figure}

We are interested in representing the sentences as hierarchical trees of dependent words. We use a freely available dependency parser named SyntaxNet\footnote{https://opensource.google.com/projects/syntaxnet}~\cite{Andor2016} to extract the dependency tree corresponding to each sentence. The child-sum TreeLSTM~\cite{Tai2015} is used to process the dependency trees. As shown in Figure~\ref{fig:model_dep_tree}, the parts-of-speech and dependency types of the words are used in addition to the GLOVE word vectors. We concatenate a parts-of-speech embedding ($\mathbf{p}_i$) and a dependency type embedding ($\mathbf{d}_i$) with the word vectors. These embeddings ($\mathbf{p}_i$ and $\mathbf{d}_i$) are learned through back-propagation along with other free parameters of the network. The mathematical description of the model is as follows:
\begin{align}
&\mathbf{x'}_t = [\mathbf{w}'_t, \mathbf{p}'_t, \mathbf{d}'_t]\label{eq:concat}\\
&\mathbf{\tilde{h}}_t = \sum_{k\in C(t)}\mathbf{h}_k\label{eq:treelstm_first}\\
&\mathbf{i}_t =\sigma(\mathbf{U}_i\mathbf{x}_t+\mathbf{V}_i\mathbf{\tilde{h}}_{t} + \mathbf{b}_i)\label{eq:Vstart}\\
&\mathbf{f}_{tk}=\sigma(\mathbf{U}_f\mathbf{x}_t+\mathbf{V}_f\mathbf{h}_k + \mathbf{b}_f)\\
&\mathbf{u}_t =\tanh(\mathbf{U}_u\mathbf{x}_t+\mathbf{V}_u\mathbf{\tilde{h}}_{t} + \mathbf{b}_u)\\
&\mathbf{o}_t =\sigma(\mathbf{U}_o\mathbf{x}_t+\mathbf{V}_o\mathbf{\tilde{h}}_t + \mathbf{b}_o)\label{eq:Vend}\\
&\mathbf{c}_t =\mathbf{f}_{tk}\odot \mathbf{c}_k + \mathbf{i}_t\odot\mathbf{u}_t\\
&\mathbf{h}_t = \mathbf{o}_t\odot\tanh(\mathbf{c}_t)\label{eq:treelstm_last} \\
&\mathbf{h}_{s_j} = \mathbf{h}_{ROOT}\\
&\mathbf{h} = \frac{1}{N}\sum_{j=1}^N\mathbf{h}_{s_j}\\
&\mathbf{r} = \sigma(\mathbf{W}\mathbf{h} + \mathbf{b}_r)
\end{align}
Here equation~\eqref{eq:concat} represents the concatenation of the pre-trained GLOVE word-vectors with the learnable embeddings for the parts of speech and the dependency type of a word. $C(t)$ represents the set of all the children of node $t$. The parent-child relation of the treeLSTM nodes come from the dependency tree. Zero vectors are used as the children of leaf nodes. A node $t$ is processed recursively using the equations~\eqref{eq:treelstm_first} through~\eqref{eq:treelstm_last}. Notably, these equations are similar to equations~\eqref{eq:lstm_first} to~\eqref{eq:lstm_last}, except the fact that the memory cell and hidden states flow hierarchically from the children to the parent instead of sequential movement. Each node contains a forget gate ($\mathbf{f}$) for each child. The sentence embedding vector is obtained from the root node.

\section{Training the Networks}
We implemented the networks in pyTorch~\footnote{pytorch.org}. Details of the training procedure are described in the following subsections.

\subsection{Optimization}\label{sec:optimization}
We use multi-label Binary Cross-Entropy loss as defined below for the backpropagation of the gradients:
\begin{equation}
\ell(\mathbf{r},\mathbf{y}) = -\frac{1}{n}\sum_{i=1}^{n}(y_i\log(r_i) + (1-y_i)\log(1-r_i))
\end{equation}
Here $\mathbf{r}$ is the model output and $\mathbf{y}$ is the ground truth label obtained from data. $r_i$ and $y_i$ represent the $i^{\text{th}}$ element of $\mathbf{r}$ and $\mathbf{y}$. $n=14$ represents the number of the rating categories.

We randomly split the training dataset into 9:1 ratio and name them training and development subsets respectively. The networks are trained over the training subset. We use the loss in the development subset to tune the hyper-parameters, to adjust the learning rate, to control the regularization strength, and to select the best model for final evaluation. The training loop is terminated when the loss over the development subset saturates. The model parameters are saved only when the loss over the development subset is lower than any previous iteration.

We experiment with two optimization algorithms: Adam~\cite{Kingma2014} and Adagrad~\cite{Duchi2011}. The learning rate is varied in an exponential range from $0.0001$ to $1$. The optimization algorithms are evaluated with mini-batches of size $10$, $30$, and $50$. We obtain the best results using Adagrad with learning rate $0.01$ and in Adam with a learning rate of $0.00066$. The training loop ran for $50$ iterations which mostly saturates the development set loss. We conducted around $100$ experiments with various parameters. Each experiment usually takes about 120 hours to make 50 iterations over the dataset when running in an Nvidia K20 GPU.
\begin{figure}
\centering
\includegraphics[width=\linewidth]{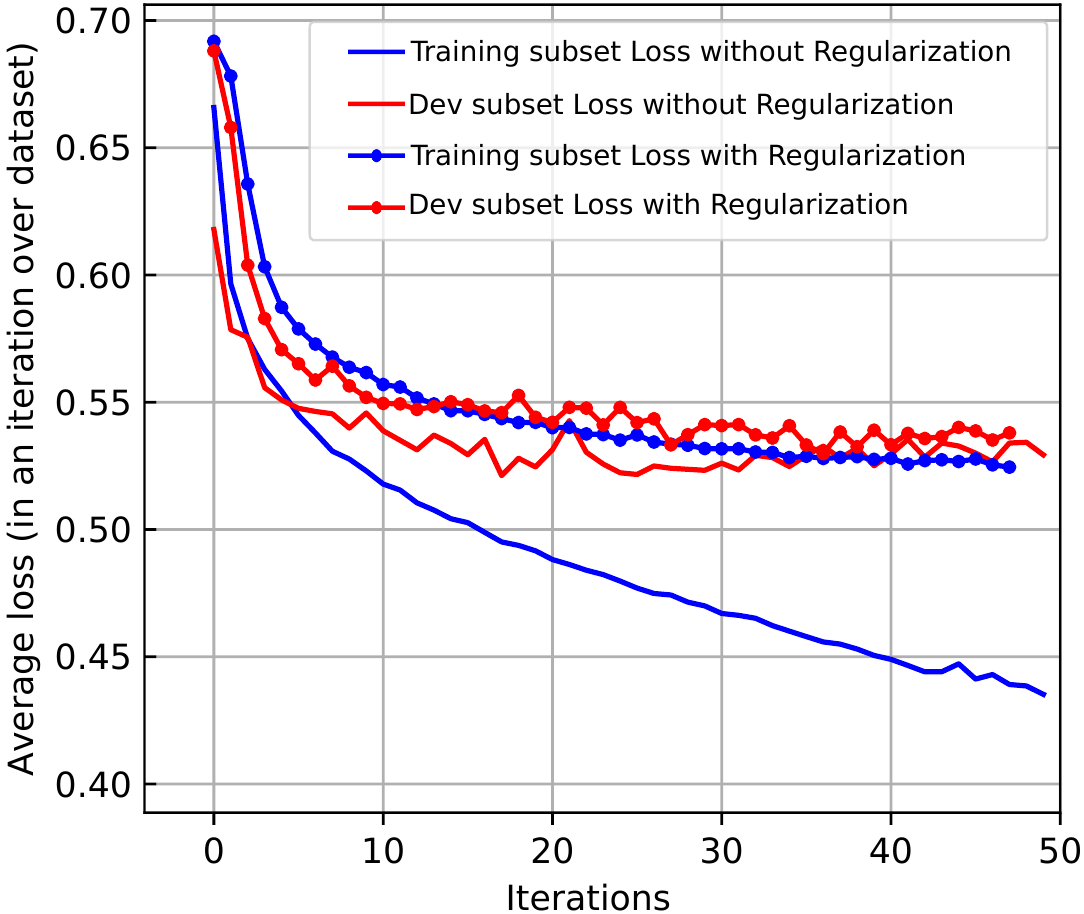}
\caption{Effect of regularization on the training and development subset loss}
\label{fig:weight-drop}
\end{figure}
\subsection{Regularization}
Neural networks are often regularized using Dropout~\cite{Hinton2012} to prevent overfitting---where the elements of a layer's output are set to zero with a probability $p$ during the training time. A naive application of dropout to LSTM's hidden state disrupts its ability to retain long-term memory. We resolve this issue using the weight-dropping technique~\cite{Wan2013,Merity2017}. In this technique, instead of applying the dropout operation between every time-steps, it is applied to the hidden-to-hidden weight matrices~\cite{Wan2013} (i.e. the $\mathbf{V}$ matrices in equations~\eqref{eq:lstm_first} to~\eqref{eq:Vmatend} and~\eqref{eq:Vstart} to~\eqref{eq:Vend}). We use the original dropout method in the fully connected layers. The dropout probability, $p$ is set to $0.2$. The effect of regularization is shown in Figure~\ref{fig:weight-drop}. We also experimented with weight-decay regularization, which adds the average $\ell2$-norm of all the network parameters to the loss function. However, weight-decay adversely affected the training process in our neural network models.

\section{Baseline Methods}
We compare the performance of the neural network models against several machine learning techniques. We also compare our results with the one reported in \citet{weninger2013words}.


We use a psycholinguistic lexicon named ``Linguist Inquiry Word Count'' (LIWC)~\cite{Pennebaker-liwc01} for extracting the language features. We count the total number of words under the 64 word categories provided in the LIWC lexicon and normalize these counts by the total number of words in the transcript. The LIWC categories include functional words (e.g., articles, quantifiers, pronouns), various content categories (e.g., anxiety, insight), positive emotions (e.g., happy, kind), negative emotions (e.g., sad, angry), and more. These features have been used in several related works~\cite{Ranganath2009,Zechner2009,Naim2016,Liu2017} with good prediction performance.

We use the Linear Support Vector Machine (SVM)~\cite{Vapnik1964} and LASSO~\cite{Tibshirani1996} as the baseline classifiers. In SVM, the following objective function is minimized:
\begin{equation}
\begin{aligned}
& \underset{\mathbf{w}, \xi_i, b}{\text{minimize}}
& & \frac{1}{2}  \|  \mathbf{w} \|^2  + C \sum_{i = 1}^N \xi_i\\
& \text{subject to}
& & y_i \left(\mathbf{w}' \mathbf{x}_i - b\right) \geq 1 - \xi_i,  \ \forall i \\
&&&  \xi_i \geq 0,  \ \forall i \\
\end{aligned}
\end{equation}
Where $\mathbf{w}$ is the weight vector and $b$ the bias term. $\|\mathbf{w}\|$ refers to the $\ell2$ norm of the vector $\mathbf{w}$. In these equations, we assume that the ``higher than median'' and ``lower than median'' classes are represented by $1$ and $-1$ values respectively.

We adapt the original LASSO~\cite{Tibshirani1996} regression model for the classification purposes. It is equivalent to Logistic regression with $\ell1$ norm regularization. It works by solving the following optimization problem:
\begin{equation}
\begin{aligned}
&\underset{\mathbf{w},b}{\text{minimize}}
\quad \| \mathbf{w} \|_1 + k\\
& k=C\sum_{i=1}^N \log\left(\exp\left(-y_i\left(\mathbf{w}' \mathbf{x}_i + b \right)\right)+1\right) \\
\end{aligned}
\end{equation}
where $C > 0$ is the inverse of the regularization strength, and  $\| \mathbf{w} \|_1 = \sum_{j=1}^d |w_j|$ is the $\ell1$ norm of $\mathbf{w}$. The $\ell1$ norm regularization is known to push the coefficients of the irrelevant features down to zero, thus reducing the predictor variance.

\begin{table}
\centering
\begin{tabular}{lrrrr} 
\toprule
 \textbf{Model}               & \textbf{\makecell{Avg.\\ F-sc.}} & \textbf{\makecell{Avg.\\Prec.}} & \textbf{\makecell{Avg.\\ Rec.}} & \textbf{\makecell{Avg.\\ Acc.}}  \\ 
\midrule
\textbf{Word Seq}             & 0.76                                     & 0.76                                     & 0.76                                   & 0.76                \\
\textbf{Dep. Tree}            & \textbf{0.77}                            & \textbf{0.77}                            & \textbf{0.77}                          & \textbf{0.77}                \\
\midrule
\textbf{\makecell[l]{Dep. Tree\\ (Unscaled)}} & 0.67                 & 0.70                 & 0.68               & 0.68                \\ 
\midrule
\textbf{LinearSVM}            & 0.69                                     & 0.69                                     & 0.69                                   & 0.69                \\
\textbf{LASSO}                & 0.69                                     & 0.70                                     & 0.70                                   & 0.70                \\ 
\midrule
\textbf{Weninger et al.}                      & -- & --                   &     0.71          & -- \\
\bottomrule
\end{tabular}
\caption{Average F-score, Precision, Recall and Accuracy for various models. Due to the choice of the median thresholds, the precision, recall, F-score, and accuracy values are practically identical in our experiments.}
\label{tab:avg_metrics}
\end{table}
\section{Experimental Results}\label{sec:exp_res}
We allocated $150$ randomly sampled TED talks from the dataset as a reserved test subset. Data from this subset was never used for training the models or for tuning the hyper-parameters. We used it only for evaluating the models saved in the training process. All the results shown in this section are computed over this test subset. 

We evaluate the predictive models by computing four performance metrics---F-score, Precision, Recall, and Accuracy. We compute the average of each metric over all the rating categories which are shown in Table~\ref{tab:avg_metrics}. The first two rows represent the average performances of the Word Sequence model and the Dependency Tree based model respectively. These models were trained and tested on the scaled rating counts ($R_\text{scaled}$). The dependency tree based model shows a slightly better performance than the word sequence model. We also trained and tested the dependency tree model with unscaled rating counts ($3^\text{rd}$ row in Table~\ref{tab:avg_metrics}). Notably, the same network architecture that performed best for the scaled ratings is now performing much worse for predicting the unscaled ratings. Modeling the data generating process and removing the effects of the unwanted variables resulted in a $10\%$ improvement in the prediction performance. Furthermore, this is achieved without the inclusion of any additional data. We believe this is because the unscaled ratings are affected by the biases present in the dataset---which are difficult to predict using the transcripts only. Therefore, removing the biases makes the prediction problem easier. We compare our results with \citet{weninger2013words} as well. The average recall for their best performing classifier (SVM) is shown in the last row of the table which is similar to our baseline methods.

\begin{table}
\centering
\begin{tabular}{lccc} 
\toprule
 \textbf{Ratings}      & \textbf{\makecell[l]{Word\\Seq.}}        & \textbf{\makecell[l]{Dep.\\ Tree}}       & \textbf{\makecell[l]{Weninger\\et al. (SVM)}}  \\ 
\midrule
\textbf{Beautiful}     & 0.88                     & \textbf{0.91}            & 0.80                     \\
\textbf{Confusing}     & 0.70                     & \textbf{0.74}            & 0.56                     \\
\textbf{Courageous}    & 0.84                     & \textbf{0.89}            & 0.79                     \\
\textbf{Fascinating}   & 0.75                     & 0.76                     & \textbf{0.80}            \\
\textbf{Funny}         & \textbf{0.78}            & 0.77                     & 0.76                     \\
\textbf{Informative}   & 0.81                     & \textbf{0.83}            & 0.78                     \\
\textbf{Ingenious}     & 0.80                     & \textbf{0.81}            & 0.74                     \\
\textbf{Inspiring}     & 0.72                     & \textbf{0.77}            & 0.72                     \\
\textbf{Jaw-dropping}  & 0.68                     & \textbf{0.72}            & \textbf{0.72}            \\
\textbf{Longwinded}    & \textbf{0.73}            & 0.70                     & 0.63                     \\
\textbf{Obnoxious}     & \textbf{0.64}            & \textbf{0.64}            & 0.61                     \\
\textbf{OK}            & \textbf{0.73}            & 0.70                     & 0.61                     \\
\textbf{Persuasive}    & 0.83                     & \textbf{0.84}            & 0.78                     \\
\textbf{Unconvincing}  & \textbf{0.70}            & \textbf{0.70}            & 0.61                     \\ 
\midrule
\textbf{Average}                & 0.76 & 0.77 & 0.71                     \\
\bottomrule
\end{tabular}
\caption{Recall for various rating categories. The reason we choose recall is for making comparison with the results reported by \citet{weninger2013words}.}
\label{tab:ratingwise_metric}
\end{table}
In Table~\ref{tab:ratingwise_metric}, we present the recall values for all the different rating categories. We choose recall over the other metrics to make a comparison with the results reported by \citet{weninger2013words}. However, all the other metrics (Accuracy, Precision, and F-score) have practically the identical value as the recall due to our choice of median threshold while preparing $R_\text{scaled}$. The highest recall is observed for the \emph{Beautiful} ratings which is $0.91$. The lowest recall is for \emph{Obnoxious}---$0.64$. We observe a trend from the table that, the ratings with fewer counts (shown in Figure~\ref{fig:rating_counts}) are usually difficult to predict.

Table~\ref{tab:ratingwise_metric} provides a clearer picture of how the dependency tree based neural network performs better than the word sequence neural network. The former achieves a higher recall for most of the rating categories ($8$ out of $14$). Only in three cases (\emph{Funny}, \emph{Longwinded}, and \emph{OK}) the word sequence model achieves higher performance than the dependency tree model. Both these models perform equally well for the \emph{Obnoxious} and \emph{Unconvincing} rating category. It is important to realize that the dependency trees we extracted from the sentences of the transcripts, were not manually annotated. They were extracted using SyntaxNet, which itself introduces some error. \citet{Andor2016} described their model accuracy to be approximately $95\%$. We expected to notice an impact of this error in the results. However, the results show that the additional information (Parts of Speech tags, Dependency Types and Structures) benefited the prediction performance despite the error in the dependency trees. We think the hierarchical tree structure resolves some ambiguities in the sentence semantics which is not available to the word sequence model.

Finally, comparison with the results from \citet{weninger2013words} reveals that the neural network models perform better for almost every rating category except \emph{Fascinating} and \emph{Jaw-Dropping}. A neural network is a universal function approximator~\cite{cybenko1989,hornik1991} and thus expected to perform better. Yet we think another reason for its excel is its ability to process a faithful representation of the transcripts. In the baseline methods, the transcripts are provided as words without any order. In the neural counterparts, however, it is possible to maintain a more natural representation of the words---either the sequence or the syntactic relationship among them through a dependency tree. Besides, neural networks intrinsically capture the correlations among the rating categories. The baseline methods, on the other hand, consider each category as a separate classification problem. These are possibly a few reasons why the neural networks are a better choice for the TED talk rating prediction task.

\section{Conclusion}
In summary, we presented neural network-based architectures to predict the TED talk ratings from the speech transcripts. We carefully modeled the data generating process from known causal relations in order to remove the effects of data bias. Our experimental results show that our method effectively removes the data bias from the prediction model. This process resulted in a $10\%$ improvement in the prediction accuracy. This result indicates that modeling the data generating process and removing the effect of unwanted variables can lead to higher predictive capacity even with a moderately sized dataset as ours. The neural network architectures provide the state of the art prediction performance, outperforming the competitive baseline method in the literature.

Our results also show that the dependency tree based neural network architecture performs better in predicting the TED talk ratings as compared to a word sequence model. The exact reason why this happens, however, remains to be explored in the future.


\bibliography{tanveer_ted_2018}
\bibliographystyle{acl_natbib}

\end{document}